\documentclass[10pt,twocolumn,letterpaper]{article}

\usepackage{cvpr}              
\usepackage[table,xcdraw]{xcolor}

\usepackage{graphicx}
\usepackage{amsmath}
\usepackage{amssymb}
\usepackage{booktabs}
\usepackage{indentfirst}
\usepackage{multicol}
\usepackage{multirow}
\usepackage{bbding}
\usepackage{color}
\usepackage{xcolor}
\usepackage{arydshln}
\usepackage{diagbox}
\usepackage{makecell}

\usepackage[pagebackref,breaklinks,colorlinks]{hyperref}

\usepackage[capitalize]{cleveref}
\crefname{section}{Sec.}{Secs.}
\Crefname{section}{Section}{Sections}
\Crefname{table}{Table}{Tables}
\crefname{table}{Tab.}{Tabs.}

\def\cvprPaperID{3} 
\def\confName{CVPRW}
\def\confYear{2023}

\begin{document}

\title{What Makes a Good Data Augmentation for Few-Shot Unsupervised Image Anomaly Detection?}

\author{Lingrui Zhang\footnotemark[1]\\
SUSTech\\
{\tt\small lingrui\_zhang@foxmail.com}
\and
Shuheng Zhang\footnotemark[1]\\
SUSTech\\
{\tt\small zhangsh2001@foxmail.com}
\and
Guoyang Xie\\
SUSTech and University of Surrey\\
{\tt\small guoyang.xie@surrey.ac.uk}
\and
Jiaqi Liu\\
SUSTech\\
{\tt\small liujq32021@mail.sustech.edu.cn}
\and
Hua Yan\\
University of Warwick\\
{\tt\small Aaron.Yan.1@warwick.ac.uk}
\and
Jinbao Wang\footnotemark[2]\\
SUSTech\\
{\tt\small linking@163.com}
\and
Feng Zheng\footnotemark[2]\\
SUSTech\\
{\tt\small zfeng02@gmail.com}
\and
Yaochu Jin\\
Bielefeld University and University of Surrey\\
{\tt\small yaochu.jin@uni-bielefeld.de}
 }






\maketitle

\footnotetext[1]{Contributed equally.}
\footnotetext[2]{Corresponding authors.}
  
\begin{abstract}
Data augmentation is a promising technique for unsupervised anomaly detection in industrial applications, where the availability of positive samples is often limited due to factors such as commercial competition and sample collection difficulties. In this paper, how to effectively select and apply data augmentation methods for unsupervised anomaly detection is studied. The impact of various data augmentation methods on different anomaly detection algorithms is systematically investigated through experiments. The experimental results show that the performance of different industrial \underline{i}mage \underline{a}nomaly \underline{d}etection (termed as IAD) algorithms is not significantly affected by the specific data augmentation method employed and that combining multiple data augmentation methods does not necessarily yield further improvements in the accuracy of anomaly detection, although it can achieve excellent results on specific methods. These findings provide useful guidance on selecting appropriate data augmentation methods for different requirements in IAD.
\end{abstract}

\section{Introduction}\label{sec:intro}
Industrial Image Anomaly Detection (IAD) is a challenging task that aims to identify defects or abnormalities in industrial images. Unlike natural images, industrial images have high similarity among normal samples, which makes it easy to model their distribution. However, the number and variety of anomalies are very low and unpredictable, which makes it hard to collect enough labeled data for supervised learning. Therefore, IAD requires unsupervised or weakly supervised methods that can learn from normal samples only and detect anomalies based on their deviation from the normal distribution. 

There are two main types of anomaly detection, feature embedding-based methods\cite{bergmann2020uninformed,salehi2021multiresolution, Wang2021StudentTeacherFP, yamada2021reconstruction, rudolph2021same, yu2021fastflow, rudolph2022fully, gudovskiy2022cflow, cohen2020sub, li2021anomaly, kim2021semi, defard2021padim, yi2020patch, hu2021semantic}, and reconstruction-based methods\cite{zavrtanik2021draem, zavrtanik2022dsr, jiang2022masked, Dehaene2020AnomalyLB, schluter2021self, zavrtanik2021reconstruction}. Feature embedding-based methods use pre-trained models to extract high-level features from industrial images. And then clustering or density estimation techniques are applied to measure the distance or probability of each image feature from the normal cluster or distribution. Images with large distances or low probability are considered to be anomalies. Reconstruction-based methods use generative models, such as variational autoencoders or generative adversarial networks, to learn a mapping function from industrial images to a latent space and vice versa. Then they reconstruct each image from its latent representation and compute the reconstruction error between the original and reconstructed images. Images with large reconstruction errors are considered to be anomalies.

Few-shot IAD is an emerging research topic\cite{huang2022registration, roth2022towards, rudolph2021same, sheynin2021hierarchical} that aims to detect defects in industrial images with only a few normal samples. This scenario is common in real-world industrial applications, where collecting a large number of normal samples may be impractical or costly. For instance, in the debugging phase of a production line, only a few normal samples can be produced before the yield reaches a satisfactory level. Moreover, some industrial domains may face commercial competition or privacy issues that prevent them from sharing large amounts of training data.

Data augmentation methods have been used in unsupervised few-shot anomaly detection\cite{schwartz2022maeday, sheynin2021hierarchical, huang2022registration, rudolph2021same} to increase the diversity and robustness of normal samples. However, there is a lack of systematic and comprehensive analysis of how different data augmentation methods affect the performance of IAD algorithms. Previous works have used various data augmentation methods with different settings, but without providing in-depth and detailed explanations or comparisons. For example, MAEDAY\cite{schwartz2022maeday} only uses rotation, while \cite{sheynin2021hierarchical} aggressively uses multiple augmentations. Only RegAD\cite{huang2022registration} is one of the few works that has studied the impact of data augmentation methods on few-shot anomaly detection in more detail, using four data augmentation methods on two IAD datasets. In this paper, we conduct a thorough and extensive study of the role of data augmentation methods. We apply 6 separate data augments to 3 IAD datasets and test them on 11 IAD algorithms. For the number of training images, we choose 1, 2, 4, and 8 respectively. In total, we conduct thousands of experiments and systematically summarize a large number of meaningful conclusions. The results of our experiments show that there is no single data augmentation method that can consistently improve the performance of all unsupervised IAD algorithms. However, we subdivide the 11 IAD methods into multiple classes, and the analysis shows that data augmentation methods tend to have similar effects on similar IAD algorithms. In addition, we explore the effects of combining different data augmentations to increase the diversity and complexity of the training data, and the experimental results show that the benefits of mixed data augmentation depend heavily on the IAD method. For most methods, we do not observe significant improvements from using multiple data augmentations. However, in some cases, we find that mixed data augmentation can boost accuracy remarkably. The main contribution of this work can be summarized as follows:
\begin{itemize}
    \item We examine 6 data augmentation methods, and 11 image anomaly detection methods on 3 benchmark datasets, resulting in a total of 6,688 instances. In addition, we present a plug-and-play and modular implementation for data augmentation of few-shot IAD evaluation, which greatly benefits the future development of few-shot IAD evaluation. 
    \item Our research highlights that the optimal choice of data augmentation method is dependent on the specific methodology employed. Interestingly, we find that comparable IAD algorithms generally respond similarly to various data augmentation methods.
    \item Our findings indicate that the effectiveness of combining multiple data augmentation methods substantially depends on the specific IAD method employed. Additionally, we identify the IAD methods that are compatible with mixed data augmentation.
\end{itemize}

\begin{figure*}[htbp]
  \centering
  \includegraphics[width=0.75\linewidth]{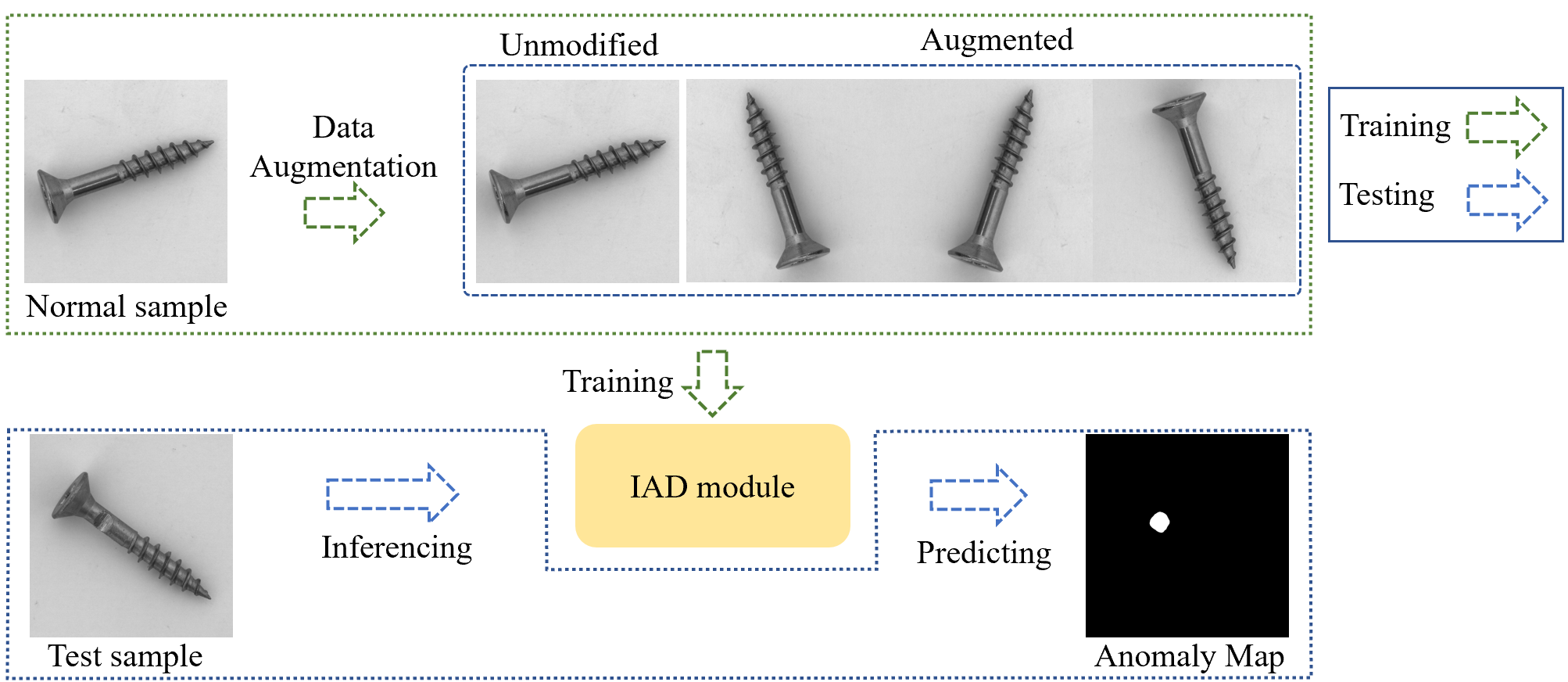}
   \caption{The overall framework of data augmentation for few-shot IAD. In training phase, the normal image and its augmented images are fed into IAD model for training. In inference phase, the test image are classified as anomalies if at least one patch is anomalous and the pixel-level anomaly segmentation is generated by the prediction result of IAD model.}
   \label{fig:framework}
\end{figure*}

\section{Related Works}
\subsection{Unsupervised IAD}
There are many kinds of anomaly detection methods, and they can be divided into two main types, embedding-based\cite{bergmann2020uninformed, rudolph2021same, cohen2020sub} and reconstruction-based\cite{yang2020dfr, zavrtanik2021draem}. The embedding-based methods are further divided into four classes, i.e., Normalizing Flow\cite{rudolph2021same}, Teacher-Student\cite{bergmann2020uninformed}, One-Class Classification\cite{sohn2020learning} and Memory Bank\cite{cohen2020sub}. Next, they will be presented in detail.

\subsubsection{Normalizing Flow (NF)}
Normalizing Flows is a generative model that can produce easy-to-handle distributions, where sampling and density assessment can be efficient and accurate\cite{rudolph2021same}. CS-Flow\cite{rudolph2022fully} used the feature block x to input the flow model to fit the distribution z of the product of Gaussian distribution and Dirac distribution.
The network is updated by the likelihood probability method of this distribution, and the evaluation index is made. 
FastFlow\cite{yu2021fastflow} is implemented by two-dimensional normalized flow and used as a probability distribution estimator, which effectively mapped image features to easily processed base distributions and pays attention to the relationship between local and global features. However, the features of normal images in industrial manufacturing may not conform to the Gaussian distribution.

\subsubsection{Teacher-Student}
It is a kind of transfer learning. That is to say, the performance of one model is transferred to another model. The teacher network is a complex network with very good performance and generalization ability. This network can be used to guide the student network to learn so that a simpler student model with less parameter computation can also have similar performance to the teacher network, which is also a way of model compression. STPM\cite{Wang2021StudentTeacherFP} is multiplied by three different resolutions of the anomaly map to detect the anomalies. RD4AD\cite{deng2022anomaly} proposed a reverse knowledge distillation scheme. It took the teacher as encoder, and the student as decoder, and added a one-class bottleneck embedding module between them. But, These methods rely too much on the teacher network, which may weaken the generalization ability.

\begin{table}[!h]
\Large
\renewcommand{\thetable}{1(a)}
\centering
\resizebox{\columnwidth}{!}{%
\begin{tabular}{l|c|ccccccc}
\hline
Shot &
  \textbf{Methods} &
  \textbf{Vanilla} &
  \textbf{Rotation} &
  \textbf{Flip} &
  \textbf{Scale} &
  \textbf{Translate} &
  \textbf{ColorJitter} &
  \textbf{Perspective} \\ \hline
 &
  \textbf{CFA} &
  0.934 &
  0.933 &
  0.93 &
  0.905 &
  0.902 &
  {\color[HTML]{FF0000} 0.937} &
  0.927 \\
 &
  \textbf{CSFlow} &
  0.678 &
  {\color[HTML]{FF0000} 0.763} &
  {\color[HTML]{00B0F0} 0.737} &
  0.679 &
  0.72 &
  0.735 &
  0.615 \\
 &
  \textbf{CutPaste} &
  0.762 &
  {\color[HTML]{FF0000} 0.893} &
  {\color[HTML]{00B0F0} 0.869} &
  0.611 &
  0.808 &
  0.831 &
  0.861 \\
 &
  \textbf{DRAEM} &
  0.665 &
  0.648 &
  {\color[HTML]{00B0F0} 0.707} &
  0.574 &
  0.672 &
  0.687 &
  {\color[HTML]{FF0000} 0.759} \\
 &
  \textbf{FastFlow} &
  0.5 &
  {\color[HTML]{FF0000} 0.868} &
  0.74 &
  {\color[HTML]{00B0F0} 0.797} &
  0.782 &
  0.787 &
  0.754 \\
 &
  \textbf{FAVAE} &
  0.713 &
  0.712 &
  {\color[HTML]{FF0000} 0.776} &
  0.662 &
  0.583 &
  0.605 &
  {\color[HTML]{00B0F0} 0.716} \\
 &
  \textbf{PaDiM} &
  0.781 &
  {\color[HTML]{FF0000} 0.879} &
  {\color[HTML]{00B0F0} 0.878} &
  0.853 &
  0.8 &
  0.811 &
  0.849 \\
 &
  \textbf{PatchCore} &
  0.906 &
  {\color[HTML]{FF0000} 0.927} &
  0.914 &
  {\color[HTML]{00B0F0} 0.92} &
  0.913 &
  0.91 &
  0.904 \\
 &
  \textbf{RD4AD} &
  0.844 &
  {\color[HTML]{00B0F0} 0.913} &
  {\color[HTML]{FF0000} 0.918} &
  0.879 &
  0.908 &
  0.887 &
  0.886 \\
 &
  \textbf{SPADE} &
  - &
  - &
  - &
  - &
  - &
  - &
  - \\
\multirow{-11}{*}{1} &
  \textbf{STPM} &
  0.876 &
  0.89 &
  {\color[HTML]{00B0F0} 0.915} &
  0.912 &
  0.888 &
  {\color[HTML]{FF0000} 0.92} &
  0.908 \\ \hline
 &
  \textbf{CFA} &
  0.928 &
  {\color[HTML]{FF0000} 0.936} &
  {\color[HTML]{00B0F0} 0.932} &
  0.916 &
  0.919 &
  0.925 &
  0.928 \\
 &
  \textbf{CSFlow} &
  0.8 &
  {\color[HTML]{00B0F0} 0.878} &
  0.76 &
  {\color[HTML]{FF0000} 0.884} &
  0.853 &
  0.833 &
  0.724 \\
 &
  \textbf{CutPaste} &
  0.718 &
  0.855 &
  {\color[HTML]{FF0000} 0.89} &
  {\color[HTML]{00B0F0} 0.883} &
  0.582 &
  0.782 &
  0.604 \\
 &
  \textbf{DRAEM} &
  0.619 &
  0.752 &
  {\color[HTML]{00B0F0} 0.772} &
  0.724 &
  0.532 &
  0.749 &
  {\color[HTML]{FF0000} 0.779} \\
 &
  \textbf{FastFlow} &
  0.644 &
  {\color[HTML]{FF0000} 0.887} &
  0.811 &
  0.844 &
  {\color[HTML]{00B0F0} 0.88} &
  0.829 &
  0.751 \\
 &
  \textbf{FAVAE} &
  0.796 &
  {\color[HTML]{00B0F0} 0.798} &
  {\color[HTML]{FF0000} 0.842} &
  0.636 &
  0.66 &
  0.65 &
  0.589 \\
 &
  \textbf{PaDiM} &
  0.875 &
  {\color[HTML]{00B0F0} 0.903} &
  {\color[HTML]{FF0000} 0.907} &
  0.878 &
  0.886 &
  0.881 &
  0.878 \\
 &
  \textbf{PatchCore} &
  0.911 &
  0.913 &
  0.91 &
  0.909 &
  {\color[HTML]{00B0F0} 0.916} &
  0.895 &
  {\color[HTML]{FF0000} 0.92} \\
 &
  \textbf{RD4AD} &
  0.885 &
  {\color[HTML]{FF0000} 0.919} &
  0.91 &
  0.911 &
  {\color[HTML]{00B0F0} 0.916} &
  0.914 &
  {\color[HTML]{FF0000} 0.919} \\
 &
  \textbf{SPADE} &
  - &
  0.882 &
  0.887 &
  0.872 &
  0.872 &
  0.831 &
  0.884 \\
\multirow{-11}{*}{2} &
  \textbf{STPM} &
  0.907 &
  0.92 &
  {\color[HTML]{00B0F0} 0.926} &
  0.924 &
  0.923 &
  {\color[HTML]{FF0000} 0.934} &
  0.923 \\ \hline
 &
  \textbf{CFA} &
  0.924 &
  {\color[HTML]{00B0F0} 0.925} &
  {\color[HTML]{FF0000} 0.926} &
  0.885 &
  0.917 &
  0.92 &
  0.909 \\
 &
  \textbf{CSFlow} &
  0.742 &
  {\color[HTML]{00B0F0} 0.814} &
  0.745 &
  {\color[HTML]{FF0000} 0.89} &
  0.756 &
  0.78 &
  0.665 \\
 &
  \textbf{CutPaste} &
  0.778 &
  {\color[HTML]{FF0000} 0.896} &
  {\color[HTML]{00B0F0} 0.891} &
  0.775 &
  0.81 &
  0.753 &
  0.608 \\
 &
  \textbf{DRAEM} &
  0.684 &
  {\color[HTML]{FF0000} 0.857} &
  0.667 &
  0.714 &
  0.632 &
  0.7 &
  {\color[HTML]{00B0F0} 0.753} \\
 &
  \textbf{FastFlow} &
  0.8 &
  0.815 &
  0.764 &
  {\color[HTML]{00B0F0} 0.824} &
  {\color[HTML]{FF0000} 0.826} &
  0.772 &
  0.749 \\
 &
  \textbf{FAVAE} &
  0.85 &
  0.714 &
  {\color[HTML]{00B0F0} 0.72} &
  0.712 &
  0.711 &
  0.513 &
  {\color[HTML]{FF0000} 0.748} \\
 &
  \textbf{PaDiM} &
  0.904 &
  0.914 &
  {\color[HTML]{FF0000} 0.923} &
  0.911 &
  {\color[HTML]{00B0F0} 0.918} &
  0.907 &
  0.912 \\
 &
  \textbf{PatchCore} &
  0.921 &
  0.919 &
  0.919 &
  {\color[HTML]{FF0000} 0.924} &
  0.919 &
  0.907 &
  0.904 \\
 &
  \textbf{RD4AD} &
  0.91 &
  0.922 &
  {\color[HTML]{FF0000} 0.927} &
  {\color[HTML]{00B0F0} 0.925} &
  0.908 &
  0.911 &
  {\color[HTML]{00B0F0} 0.925} \\
 &
  \textbf{SPADE} &
  - &
  0.871 &
  0.87 &
  0.873 &
  0.871 &
  0.84 &
  0.869 \\
\multirow{-11}{*}{4} &
  \textbf{STPM} &
  0.924 &
  0.907 &
  {\color[HTML]{00B0F0} 0.927} &
  0.923 &
  0.923 &
  {\color[HTML]{FF0000} 0.932} &
  0.914 \\ \hline
 &
  \textbf{CFA} &
  0.932 &
  {\color[HTML]{FF0000} 0.937} &
  0.924 &
  0.886 &
  0.925 &
  0.923 &
  0.923 \\
 &
  \textbf{CSFlow} &
  0.892 &
  0.884 &
  0.895 &
  0.924 &
  {\color[HTML]{00B0F0} 0.932} &
  {\color[HTML]{FF0000} 0.943} &
  0.877 \\
 &
  \textbf{CutPaste} &
  0.876 &
  {\color[HTML]{00B0F0} 0.883} &
  {\color[HTML]{FF0000} 0.904} &
  0.835 &
  0.864 &
  0.705 &
  0.755 \\
 &
  \textbf{DRAEM} &
  0.788 &
  {\color[HTML]{FF0000} 0.859} &
  {\color[HTML]{00B0F0} 0.816} &
  0.812 &
  0.761 &
  0.749 &
  0.71 \\
 &
  \textbf{FastFlow} &
  0.895 &
  0.882 &
  0.869 &
  0.885 &
  0.86 &
  0.789 &
  {\color[HTML]{FF0000} 0.896} \\
 &
  \textbf{FAVAE} &
  0.701 &
  {\color[HTML]{FF0000} 0.819} &
  {\color[HTML]{00B0F0} 0.813} &
  0.644 &
  0.606 &
  0.688 &
  0.756 \\
 &
  \textbf{PaDiM} &
  0.927 &
  {\color[HTML]{00B0F0} 0.936} &
  {\color[HTML]{00B0F0} 0.936} &
  0.926 &
  {\color[HTML]{FF0000} 0.939} &
  0.924 &
  0.931 \\
 &
  \textbf{PatchCore} &
  0.917 &
  {\color[HTML]{FF0000} 0.924} &
  {\color[HTML]{00B0F0} 0.923} &
  0.907 &
  0.912 &
  0.904 &
  0.92 \\
 &
  \textbf{RD4AD} &
  0.921 &
  {\color[HTML]{FF0000} 0.935} &
  0.93 &
  {\color[HTML]{00B0F0} 0.932} &
  0.924 &
  0.924 &
  0.931 \\
 &
  \textbf{SPADE} &
  0.893 &
  {\color[HTML]{00B0F0} 0.896} &
  0.9 &
  0.893 &
  0.895 &
  0.867 &
  {\color[HTML]{FF0000} 0.899} \\
\multirow{-11}{*}{8} &
  \textbf{STPM} &
  0.94 &
  0.897 &
  0.934 &
  0.923 &
  0.912 &
  0.928 &
  0.919 \\ \hline
\end{tabular}%
}
\caption{Image-level AUC-ROC on BTAD. The red one is the best augmentation for the IAD method, and the blue is the second best one.}
\label{tab:btad}
\end{table}

\begin{table}[!h]
\Large
\renewcommand{\thetable}{1(b)}
\centering
\resizebox{\columnwidth}{!}{%
\begin{tabular}{l|c|ccccccc}
\hline
Shot &
  \textbf{Methods} &
  \textbf{Vanilla} &
  \textbf{Rotation} &
  \textbf{Flip} &
  \textbf{Scale} &
  \textbf{Translate} &
  \textbf{ColorJitter} &
  \textbf{Perspective} \\ \hline
 &
  \textbf{CFA} &
  0.811 &
  {\color[HTML]{FF0000} 0.829} &
  0.811 &
  0.788 &
  0.802 &
  {\color[HTML]{FF0000} 0.814} &
  0.806 \\
 &
  \textbf{CSFlow} &
  0.708 &
  {\color[HTML]{FF0000} 0.727} &
  {\color[HTML]{00B0F0} 0.708} &
  {\color[HTML]{00B0F0} 0.742} &
  {\color[HTML]{FF0000} 0.75} &
  0.7 &
  0.713 \\
 &
  \textbf{CutPaste} &
  0.65 &
  {\color[HTML]{FF0000} 0.701} &
  {\color[HTML]{00B0F0} 0.702} &
  0.68 &
  0.679 &
  0.652 &
  {\color[HTML]{FF0000} 0.703} \\
 &
  \textbf{DRAEM} &
  0.683 &
  {\color[HTML]{00B0F0} 0.718} &
  {\color[HTML]{00B0F0} 0.715} &
  0.714 &
  0.69 &
  0.687 &
  {\color[HTML]{FF0000} 0.741} \\
 &
  \textbf{FastFlow} &
  0.527 &
  {\color[HTML]{FF0000} 0.618} &
  0.613 &
  {\color[HTML]{FF0000} 0.694} &
  {\color[HTML]{00B0F0} 0.682} &
  0.578 &
  0.6 \\
 &
  \textbf{FAVAE} &
  0.651 &
  0.56 &
  {\color[HTML]{FF0000} 0.591} &
  0.6 &
  0.581 &
  0.588 &
  {\color[HTML]{00B0F0} 0.626} \\
 &
  \textbf{PaDiM} &
  0.684 &
  {\color[HTML]{FF0000} 0.697} &
  {\color[HTML]{00B0F0} 0.683} &
  0.669 &
  0.683 &
  0.681 &
  0.674 \\
 &
  \textbf{PatchCore} &
  0.788 &
  {\color[HTML]{FF0000} 0.805} &
  0.792 &
  {\color[HTML]{00B0F0} 0.788} &
  {\color[HTML]{00B0F0} 0.8} &
  0.797 &
  0.789 \\
 &
  \textbf{RD4AD} &
  0.77 &
  {\color[HTML]{00B0F0} 0.805} &
  {\color[HTML]{FF0000} 0.802} &
  0.799 &
  {\color[HTML]{FF0000} 0.823} &
  {\color[HTML]{00B0F0} 0.816} &
  0.784 \\
 &
  \textbf{SPADE} &
  - &
  - &
  - &
  - &
  - &
  - &
  - \\
\multirow{-11}{*}{1} &
  \textbf{STPM} &
  0.799 &
  {\color[HTML]{00B0F0} 0.841} &
  {\color[HTML]{00B0F0} 0.814} &
  0.823 &
  {\color[HTML]{FF0000} 0.843} &
  {\color[HTML]{FF0000} 0.831} &
  0.84 \\ \hline
 &
  \textbf{CFA} &
  0.839 &
  {\color[HTML]{FF0000} 0.86} &
  {\color[HTML]{00B0F0} 0.853} &
  0.795 &
  0.825 &
  0.833 &
  0.843 \\
 &
  \textbf{CSFlow} &
  0.745 &
  {\color[HTML]{00B0F0} 0.781} &
  0.773 &
  {\color[HTML]{FF0000} 0.783} &
  0.768 &
  0.754 &
  0.778 \\
 &
  \textbf{CutPaste} &
  0.697 &
  0.709 &
  {\color[HTML]{FF0000} 0.748} &
  {\color[HTML]{00B0F0} 0.659} &
  0.659 &
  0.611 &
  {\color[HTML]{00B0F0} 0.726} \\
 &
  \textbf{DRAEM} &
  0.78 &
  0.765 &
  {\color[HTML]{00B0F0} 0.774} &
  0.751 &
  0.771 &
  0.773 &
  {\color[HTML]{FF0000} 0.784} \\
 &
  \textbf{FastFlow} &
  0.555 &
  {\color[HTML]{00B0F0} 0.743} &
  0.674 &
  {\color[HTML]{FF0000} 0.753} &
  {\color[HTML]{00B0F0} 0.731} &
  0.628 &
  0.617 \\
 &
  \textbf{FAVAE} &
  0.667 &
  {\color[HTML]{00B0F0} 0.648} &
  {\color[HTML]{FF0000} 0.659} &
  0.595 &
  0.62 &
  0.642 &
  0.631 \\
 &
  \textbf{PaDiM} &
  0.708 &
  {\color[HTML]{FF0000} 0.734} &
  {\color[HTML]{00B0F0} 0.731} &
  0.694 &
  0.716 &
  0.71 &
  0.714 \\
 &
  \textbf{PatchCore} &
  0.795 &
  {\color[HTML]{FF0000} 0.831} &
  0.805 &
  0.796 &
  {\color[HTML]{00B0F0} 0.806} &
  0.793 &
  {\color[HTML]{FF0000} 0.802} \\
 &
  \textbf{RD4AD} &
  0.798 &
  {\color[HTML]{00B0F0} 0.832} &
  0.816 &
  {\color[HTML]{FF0000} 0.835} &
  {\color[HTML]{00B0F0} 0.817} &
  {\color[HTML]{FF0000} 0.835} &
  {\color[HTML]{FF0000} 0.82} \\
 &
  \textbf{SPADE} &
  - &
  0.737 &
  0.731 &
  0.71 &
  0.734 &
  0.733 &
  0.744 \\
\multirow{-11}{*}{2} &
  \textbf{STPM} &
  0.84 &
  0.839 &
  {\color[HTML]{00B0F0} 0.85} &
  0.848 &
  {\color[HTML]{FF0000} 0.851} &
  {\color[HTML]{FF0000} 0.849} &
  0.838 \\ \hline
 &
  \textbf{CFA} &
  0.879 &
  {\color[HTML]{00B0F0} 0.891} &
  {\color[HTML]{FF0000} 0.861} &
  0.818 &
  0.86 &
  {\color[HTML]{FF0000} 0.895} &
  0.883 \\
 &
  \textbf{CSFlow} &
  0.785 &
  {\color[HTML]{FF0000} 0.841} &
  0.815 &
  {\color[HTML]{00B0F0} 0.825} &
  0.81 &
  0.806 &
  0.738 \\
 &
  \textbf{CutPaste} &
  0.728 &
  {\color[HTML]{FF0000} 0.771} &
  {\color[HTML]{00B0F0} 0.714} &
  0.519 &
  0.674 &
  0.621 &
  0.671 \\
 &
  \textbf{DRAEM} &
  0.82 &
  {\color[HTML]{FF0000} 0.798} &
  0.802 &
  0.819 &
  {\color[HTML]{00B0F0} 0.824} &
  0.823 &
  {\color[HTML]{FF0000} 0.826} \\
 &
  \textbf{FastFlow} &
  0.693 &
  0.741 &
  0.745 &
  {\color[HTML]{FF0000} 0.79} &
  {\color[HTML]{00B0F0} 0.782} &
  0.723 &
  0.734 \\
 &
  \textbf{FAVAE} &
  0.655 &
  0.626 &
  {\color[HTML]{00B0F0} 0.669} &
  0.605 &
  0.639 &
  {\color[HTML]{FF0000} 0.67} &
  {\color[HTML]{FF0000} 0.623} \\
 &
  \textbf{PaDiM} &
  0.722 &
  {\color[HTML]{FF0000} 0.734} &
  {\color[HTML]{FF0000} 0.723} &
  0.689 &
  {\color[HTML]{00B0F0} 0.72} &
  {\color[HTML]{00B0F0} 0.727} &
  0.716 \\
 &
  \textbf{PatchCore} &
  0.844 &
  {\color[HTML]{FF0000} 0.872} &
  0.826 &
  {\color[HTML]{FF0000} 0.844} &
  0.848 &
  {\color[HTML]{00B0F0} 0.852} &
  0.847 \\
 &
  \textbf{RD4AD} &
  0.838 &
  {\color[HTML]{FF0000} 0.897} &
  {\color[HTML]{FF0000} 0.871} &
  {\color[HTML]{00B0F0} 0.866} &
  {\color[HTML]{00B0F0} 0.876} &
  0.872 &
  {\color[HTML]{00B0F0} 0.861} \\
 &
  \textbf{SPADE} &
  - &
  0.764 &
  0.739 &
  0.749 &
  0.758 &
  0.757 &
  0.759 \\
\multirow{-11}{*}{4} &
  \textbf{STPM} &
  0.864 &
  0.869 &
  {\color[HTML]{00B0F0} 0.865} &
  {\color[HTML]{00B0F0} 0.876} &
  0.875 &
  {\color[HTML]{FF0000} 0.868} &
  {\color[HTML]{FF0000} 0.885} \\ \hline
 &
  \textbf{CFA} &
  0.923 &
  {\color[HTML]{FF0000} 0.913} &
  0.888 &
  0.875 &
  0.92 &
  0.919 &
  {\color[HTML]{FF0000} 0.924} \\
 &
  \textbf{CSFlow} &
  0.856 &
  {\color[HTML]{FF0000} 0.9} &
  0.863 &
  {\color[HTML]{FF0000} 0.9} &
  {\color[HTML]{00B0F0} 0.893} &
  {\color[HTML]{FF0000} 0.892} &
  0.861 \\
 &
  \textbf{CutPaste} &
  0.705 &
  {\color[HTML]{00B0F0} 0.808} &
  {\color[HTML]{FF0000} 0.857} &
  0.473 &
  0.694 &
  0.607 &
  0.673 \\
 &
  \textbf{DRAEM} &
  0.872 &
  {\color[HTML]{FF0000} 0.892} &
  {\color[HTML]{00B0F0} 0.905} &
  0.892 &
  0.9 &
  0.904 &
  {\color[HTML]{FF0000} 0.919} \\
 &
  \textbf{FastFlow} &
  0.828 &
  0.809 &
  0.793 &
  {\color[HTML]{FF0000} 0.843} &
  0.826 &
  0.809 &
  {\color[HTML]{FF0000} 0.81} \\
 &
  \textbf{FAVAE} &
  0.701 &
  {\color[HTML]{FF0000} 0.651} &
  {\color[HTML]{00B0F0} 0.67} &
  0.619 &
  0.611 &
  0.68 &
  0.668 \\
 &
  \textbf{PaDiM} &
  0.776 &
  {\color[HTML]{FF0000} 0.81} &
  {\color[HTML]{00B0F0} 0.739} &
  0.734 &
  {\color[HTML]{FF0000} 0.761} &
  {\color[HTML]{00B0F0} 0.779} &
  0.771 \\
 &
  \textbf{PatchCore} &
  0.891 &
  {\color[HTML]{FF0000} 0.916} &
  {\color[HTML]{00B0F0} 0.876} &
  0.889 &
  {\color[HTML]{00B0F0} 0.898} &
  0.883 &
  0.891 \\
 &
  \textbf{RD4AD} &
  0.903 &
  {\color[HTML]{FF0000} 0.933} &
  0.911 &
  {\color[HTML]{00B0F0} 0.917} &
  {\color[HTML]{00B0F0} 0.921} &
  0.912 &
  0.916 \\
 &
  \textbf{SPADE} &
  0.781 &
  {\color[HTML]{FF0000} 0.794} &
  0.77 &
  0.783 &
  {\color[HTML]{00B0F0} 0.791} &
  0.788 &
  {\color[HTML]{FF0000} 0.787} \\
\multirow{-11}{*}{8} &
  \textbf{STPM} &
  0.865 &
  0.896 &
  0.87 &
  0.91 &
  {\color[HTML]{FF0000} 0.914} &
  {\color[HTML]{00B0F0} 0.912} &
  0.904 \\ \hline
\end{tabular}%
}
\caption{Image-level AUC-ROC on MVTec AD. The red one is the best augmentation for the IAD method, and the blue is the second best one.}
\label{tab:mvtec}
\end{table}

\begin{table}[!h]
\Large
\renewcommand{\thetable}{1(c)}
\centering
\resizebox{\columnwidth}{!}{%
\begin{tabular}{l|c|ccccccc}
\hline
Shot &
  \textbf{Methods} &
  \textbf{Vanilla} &
  \textbf{Rotation} &
  \textbf{Flip} &
  \textbf{Scale} &
  \textbf{Translate} &
  \textbf{ColorJitter} &
  \textbf{Perspective} \\ \hline
 &
  \textbf{CFA} &
  0.721 &
  {\color[HTML]{00B0F0} 0.76} &
  {\color[HTML]{FF0000} 0.779} &
  0.75 &
  0.729 &
  {\color[HTML]{FF0000} 0.752} &
  0.744 \\
 &
  \textbf{CSFlow} &
  0.494 &
  {\color[HTML]{FF0000} 0.454} &
  {\color[HTML]{00B0F0} 0.491} &
  {\color[HTML]{00B0F0} 0.515} &
  {\color[HTML]{00B0F0} 0.528} &
  0.473 &
  {\color[HTML]{FF0000} 0.566} \\
 &
  \textbf{CutPaste} &
  0.651 &
  {\color[HTML]{00B0F0} 0.719} &
  {\color[HTML]{FF0000} 0.732} &
  0.601 &
  0.627 &
  0.677 &
  {\color[HTML]{FF0000} 0.606} \\
 &
  \textbf{DRAEM} &
  0.522 &
  {\color[HTML]{00B0F0} 0.505} &
  {\color[HTML]{00B0F0} 0.523} &
  {\color[HTML]{FF0000} 0.525} &
  0.522 &
  0.512 &
  {\color[HTML]{FF0000} 0.507} \\
 &
  \textbf{FastFlow} &
  0.5 &
  {\color[HTML]{FF0000} 0.507} &
  0.533 &
  {\color[HTML]{FF0000} 0.559} &
  {\color[HTML]{00B0F0} 0.566} &
  0.56 &
  0.488 \\
 &
  \textbf{FAVAE} &
  0.512 &
  0.479 &
  {\color[HTML]{FF0000} 0.508} &
  {\color[HTML]{00B0F0} 0.535} &
  0.502 &
  {\color[HTML]{FF0000} 0.57} &
  {\color[HTML]{00B0F0} 0.502} \\
 &
  \textbf{PaDiM} &
  0.518 &
  {\color[HTML]{FF0000} 0.677} &
  {\color[HTML]{00B0F0} 0.676} &
  0.598 &
  0.615 &
  0.604 &
  0.616 \\
 &
  \textbf{PatchCore} &
  0.664 &
  {\color[HTML]{FF0000} 0.581} &
  {\color[HTML]{FF0000} 0.682} &
  {\color[HTML]{00B0F0} 0.67} &
  {\color[HTML]{00B0F0} 0.648} &
  0.635 &
  {\color[HTML]{00B0F0} 0.67} \\
 &
  \textbf{RD4AD} &
  0.791 &
  {\color[HTML]{00B0F0} 0.77} &
  {\color[HTML]{FF0000} 0.834} &
  0.722 &
  {\color[HTML]{FF0000} 0.749} &
  {\color[HTML]{00B0F0} 0.716} &
  0.746 \\
 &
  \textbf{SPADE} &
  - &
  - &
  - &
  - &
  - &
  - &
  - \\
\multirow{-11}{*}{1} &
  \textbf{STPM} &
  0.561 &
  {\color[HTML]{00B0F0} 0.589} &
  {\color[HTML]{00B0F0} 0.585} &
  {\color[HTML]{00B0F0} 0.706} &
  {\color[HTML]{FF0000} 0.642} &
  {\color[HTML]{FF0000} 0.773} &
  0.662 \\ \hline
 &
  \textbf{CFA} &
  0.729 &
  {\color[HTML]{FF0000} 0.755} &
  {\color[HTML]{FF0000} 0.793} &
  0.744 &
  0.745 &
  {\color[HTML]{00B0F0} 0.759} &
  0.748 \\
 &
  \textbf{CSFlow} &
  0.513 &
  {\color[HTML]{00B0F0} 0.469} &
  0.515 &
  {\color[HTML]{FF0000} 0.524} &
  {\color[HTML]{FF0000} 0.548} &
  0.514 &
  {\color[HTML]{00B0F0} 0.547} \\
 &
  \textbf{CutPaste} &
  0.58 &
  0.51 &
  {\color[HTML]{FF0000} 0.757} &
  {\color[HTML]{00B0F0} 0.657} &
  {\color[HTML]{00B0F0} 0.696} &
  0.651 &
  {\color[HTML]{00B0F0} 0.66} \\
 &
  \textbf{DRAEM} &
  0.517 &
  0.528 &
  {\color[HTML]{00B0F0} 0.53} &
  0.503 &
  0.499 &
  0.499 &
  {\color[HTML]{FF0000} 0.555} \\
 &
  \textbf{FastFlow} &
  0.613 &
  {\color[HTML]{00B0F0} 0.52} &
  0.529 &
  {\color[HTML]{FF0000} 0.58} &
  {\color[HTML]{00B0F0} 0.545} &
  {\color[HTML]{FF0000} 0.638} &
  0.543 \\
 &
  \textbf{FAVAE} &
  0.51 &
  {\color[HTML]{00B0F0} 0.528} &
  {\color[HTML]{FF0000} 0.51} &
  0.512 &
  0.505 &
  {\color[HTML]{FF0000} 0.569} &
  0.518 \\
 &
  \textbf{PaDiM} &
  0.63 &
  {\color[HTML]{00B0F0} 0.706} &
  {\color[HTML]{FF0000} 0.709} &
  0.63 &
  0.658 &
  0.665 &
  0.646 \\
 &
  \textbf{PatchCore} &
  0.632 &
  {\color[HTML]{FF0000} 0.644} &
  0.649 &
  {\color[HTML]{00B0F0} 0.671} &
  {\color[HTML]{FF0000} 0.679} &
  0.67 &
  {\color[HTML]{FF0000} 0.624} \\
 &
  \textbf{RD4AD} &
  0.574 &
  {\color[HTML]{00B0F0} 0.768} &
  {\color[HTML]{FF0000} 0.845} &
  {\color[HTML]{FF0000} 0.761} &
  {\color[HTML]{00B0F0} 0.765} &
  {\color[HTML]{FF0000} 0.752} &
  {\color[HTML]{FF0000} 0.764} \\
 &
  \textbf{SPADE} &
  - &
  0.687 &
  0.695 &
  0.698 &
  0.696 &
  0.717 &
  0.678 \\
\multirow{-11}{*}{2} &
  \textbf{STPM} &
  0.62 &
  0.658 &
  {\color[HTML]{00B0F0} 0.645} &
  {\color[HTML]{00B0F0} 0.749} &
  {\color[HTML]{FF0000} 0.743} &
  {\color[HTML]{FF0000} 0.756} &
  0.715 \\ \hline
 &
  \textbf{CFA} &
  0.804 &
  {\color[HTML]{00B0F0} 0.783} &
  {\color[HTML]{FF0000} 0.804} &
  0.717 &
  0.779 &
  {\color[HTML]{FF0000} 0.803} &
  0.767 \\
 &
  \textbf{CSFlow} &
  0.504 &
  {\color[HTML]{FF0000} 0.43} &
  0.508 &
  {\color[HTML]{00B0F0} 0.552} &
  0.522 &
  0.496 &
  {\color[HTML]{FF0000} 0.559} \\
 &
  \textbf{CutPaste} &
  0.609 &
  {\color[HTML]{00B0F0} 0.768} &
  {\color[HTML]{00B0F0} 0.738} &
  0.641 &
  0.449 &
  0.639 &
  {\color[HTML]{FF0000} 0.795} \\
 &
  \textbf{DRAEM} &
  0.537 &
  {\color[HTML]{00B0F0} 0.588} &
  0.554 &
  0.553 &
  {\color[HTML]{00B0F0} 0.546} &
  0.522 &
  {\color[HTML]{FF0000} 0.62} \\
 &
  \textbf{FastFlow} &
  0.504 &
  0.584 &
  {\color[HTML]{FF0000} 0.763} &
  {\color[HTML]{FF0000} 0.499} &
  {\color[HTML]{00B0F0} 0.658} &
  0.621 &
  0.553 \\
 &
  \textbf{FAVAE} &
  0.536 &
  0.547 &
  {\color[HTML]{00B0F0} 0.557} &
  {\color[HTML]{FF0000} 0.565} &
  0.525 &
  {\color[HTML]{00B0F0} 0.559} &
  {\color[HTML]{FF0000} 0.521} \\
 &
  \textbf{PaDiM} &
  0.718 &
  {\color[HTML]{FF0000} 0.755} &
  {\color[HTML]{00B0F0} 0.751} &
  0.708 &
  {\color[HTML]{00B0F0} 0.73} &
  {\color[HTML]{00B0F0} 0.744} &
  0.713 \\
 &
  \textbf{PatchCore} &
  0.676 &
  {\color[HTML]{FF0000} 0.649} &
  {\color[HTML]{00B0F0} 0.694} &
  {\color[HTML]{FF0000} 0.698} &
  0.677 &
  {\color[HTML]{00B0F0} 0.668} &
  0.68 \\
 &
  \textbf{RD4AD} &
  0.815 &
  {\color[HTML]{FF0000} 0.8} &
  {\color[HTML]{FF0000} 0.828} &
  {\color[HTML]{00B0F0} 0.78} &
  {\color[HTML]{00B0F0} 0.808} &
  0.805 &
  {\color[HTML]{00B0F0} 0.794} \\
 &
  \textbf{SPADE} &
  - &
  0.686 &
  0.689 &
  0.695 &
  0.696 &
  0.706 &
  0.652 \\
\multirow{-11}{*}{4} &
  \textbf{STPM} &
  0.691 &
  0.697 &
  {\color[HTML]{00B0F0} 0.647} &
  {\color[HTML]{00B0F0} 0.751} &
  0.748 &
  {\color[HTML]{FF0000} 0.765} &
  {\color[HTML]{FF0000} 0.745} \\ \hline
 &
  \textbf{CFA} &
  0.821 &
  {\color[HTML]{FF0000} 0.803} &
  {\color[HTML]{00B0F0} 0.826} &
  0.756 &
  0.78 &
  0.822 &
  {\color[HTML]{FF0000} 0.832} \\
 &
  \textbf{CSFlow} &
  0.678 &
  {\color[HTML]{FF0000} 0.695} &
  0.721 &
  {\color[HTML]{FF0000} 0.72} &
  {\color[HTML]{FF0000} 0.758} &
  {\color[HTML]{00B0F0} 0.749} &
  0.675 \\
 &
  \textbf{CutPaste} &
  0.747 &
  {\color[HTML]{FF0000} 0.787} &
  {\color[HTML]{FF0000} 0.734} &
  0.545 &
  0.594 &
  0.552 &
  0.492 \\
 &
  \textbf{DRAEM} &
  0.637 &
  {\color[HTML]{00B0F0} 0.715} &
  {\color[HTML]{00B0F0} 0.698} &
  0.559 &
  0.635 &
  0.664 &
  {\color[HTML]{FF0000} 0.726} \\
 &
  \textbf{FastFlow} &
  0.544 &
  0.587 &
  {\color[HTML]{FF0000} 0.75} &
  {\color[HTML]{FF0000} 0.512} &
  0.579 &
  {\color[HTML]{00B0F0} 0.665} &
  {\color[HTML]{FF0000} 0.565} \\
 &
  \textbf{FAVAE} &
  0.566 &
  {\color[HTML]{FF0000} 0.556} &
  {\color[HTML]{00B0F0} 0.547} &
  0.499 &
  0.527 &
  0.557 &
  0.541 \\
 &
  \textbf{PaDiM} &
  0.766 &
  {\color[HTML]{00B0F0} 0.779} &
  {\color[HTML]{FF0000} 0.782} &
  0.735 &
  {\color[HTML]{FF0000} 0.759} &
  {\color[HTML]{00B0F0} 0.779} &
  0.752 \\
 &
  \textbf{PatchCore} &
  0.711 &
  {\color[HTML]{FF0000} 0.666} &
  {\color[HTML]{00B0F0} 0.693} &
  0.662 &
  {\color[HTML]{FF0000} 0.729} &
  0.69 &
  0.71 \\
 &
  \textbf{RD4AD} &
  0.825 &
  {\color[HTML]{FF0000} 0.806} &
  0.824 &
  {\color[HTML]{00B0F0} 0.798} &
  {\color[HTML]{00B0F0} 0.819} &
  0.797 &
  0.809 \\
 &
  \textbf{SPADE} &
  0.705 &
  {\color[HTML]{FF0000} 0.72} &
  {\color[HTML]{FF0000} 0.726} &
  0.706 &
  {\color[HTML]{00B0F0} 0.723} &
  0.72 &
  {\color[HTML]{FF0000} 0.713} \\
\multirow{-11}{*}{8} &
  \textbf{STPM} &
  0.732 &
  0.619 &
  0.733 &
  {\color[HTML]{FF0000} 0.769} &
  {\color[HTML]{00B0F0} 0.747} &
  {\color[HTML]{00B0F0} 0.718} &
  0.727 \\ \hline
\end{tabular}%
}
\caption{Image-level AUC-ROC on MTD. The red one is the best augmentation for the IAD method, and the blue is the second best one.}
\label{tab:mtd}
\end{table}

\subsubsection{One-Class Classification (OCC)}
Since we only need to distinguish between abnormal and normal, we can turn the problem into OCC. CutPaste\cite{li2021cutpaste} proposed a high-performance image defect anomaly detection model that can detect anomaly patterns without relying on abnormal data. The whole framework can be described in two-stage. Firstly, the representation of normal images is learned by a self-supervised learning method. Then, a single classifier is constructed based on the learned image representation. It mainly constructs negative samples by cutting images and then pasting them to other locations.

\subsubsection{Memory Bank}
This is a simple and effective method. In the training phase, the feature extraction of the normal image is stored in the memory bank. In the test phase, the distance between the feature of the test sample and the nearest neighbor feature is calculated to achieve anomaly detection. CFA\cite{lee2022cfa} proposed a new method to obtain discriminative features through metric learning. PaDiM\cite{defard2021padim} proposed an anomaly detection and localization framework based on distributed modeling in the one-class learning environment. In order to improve the effectiveness of features, SPADE\cite{cohen2020sub} extracted multi-resolution features in pyramid architecture. PatchCore\cite{roth2022towards} extended the work of a series of unsupervised anomaly detection algorithms such as SPADE, and PaDiM. It mainly solved the problem of the slow testing speed of SPADE and did some exploration in the feature extraction part.

\subsubsection{Reconstruction-Based Method}
This method is mainly to reconstruct the abnormal image into a normal image and compared it with the original abnormal image to find the abnormal part. The method proposed by DRAEM\cite{zavrtanik2021draem} provided an arbitrary number of abnormal samples and a pixel-perfect abnormal segmentation map, which could be used to train the proposed method without real abnormal samples. This method learned the joint representation of abnormal images and their non-abnormal reconstruction and learned the decision bounds of normal and abnormal samples. This method could directly locate anomalies without additional complex post-processing of network output and can be trained using simple and general anomaly simulations. Unlike DRAEM, FAVAE\cite{Dehaene2020AnomalyLB} did not rely on external data sets to generate anomalies to further enhance the generalization of the model.

\subsection{Few-Shot IAD}
Few-shot anomaly detection (FSAD) emerged not long ago. The number of samples is small, and the amount of data is not much. For how to use the data to get a better effect, RegAD\cite{huang2022registration} proposed a comparison-based solution, which is very different from the popular methods based on reconstruction or single classification. And \cite{rudolph2021same} also used 16 images to train its model. Its experimental results on the challenging and newly proposed MVTec AD datasets and Magnetic Tile Defects datasets showed that their method outperforms the existing methods. However, no one has systematically explored the field of Few-Shot IAD.

\subsection{Data Augmentation}
Data augmentation plays a great role in few-shot training. It can help the model to learn more features from data to get a better effect. According to \cite{yang2022image}, image augmentation methods based on image erasure usually delete one or more sub-regions in the image to replace the pixel values of these sub-regions with constant or random values. And image mixed data augmentation is mainly done by mixing multiple images or sub-regions of the image into one. But, these data augmentation methods cannot be used for anomaly detection of industrial products.

\section{Benchmark Setup}
Figure \ref{fig:framework} illustrates our data augmentation process for training the anomaly detection model. We apply different data augmentation to one image to generate multiple augmented images. These images are used to train the model to learn the normal appearance and features of the data. Then, we can feed a new image to the trained model and obtain two outputs: a confidence score that indicates how likely the image is anomalous, and a mask that highlights the anomalous regions in the image.

\subsection{Dataset}
We use a number of datasets to investigate the effects of different methods and models on few-shot, including MVtec AD\cite{Bergmann2019MVTecA},  MTD\cite{huang2020surface} and BTAD\cite{mishra2021vt}.

\textbf{MVTec AD}\cite{Bergmann2019MVTecA} contains five textures and ten objects in different fields. It has 4096 normal images and 1258  abnormal images with resolutions varying from 700 × 700 to 1024 × 1024.

\textbf{MTD}\cite{huang2020surface} has 952 normal images and 392 abnormal images. It contains 6 types of anomalies and simulates authenticity.

\textbf{BTAD}\cite{mishra2021vt} contains 2250 normal images and 580 abnormal images. It is a real-world dataset used for anomaly detection, and it has three industrial products with surface flaws.

\subsection{Baseline Methods}
We conduct experiments on a total of eleven methods, which are divided into two categories: feature-embedding-based and reconstruction-based methods. As mentioned earlier, feature-embedding-based methods are further divided into four classes. CSFlow\cite{rudolph2022fully} and FastFlow\cite{yu2021fastflow} are methods based on normalizing flow, and CFA\cite{lee2022cfa}, SPADE\cite{cohen2020sub}, PaDiM\cite{defard2021padim}, and PatchCore\cite{roth2022towards} all use the features of good samples saved by memory bank. In addition, both RD4AD\cite{deng2022anomaly} and STPM\cite{Wang2021StudentTeacherFP} use the teacher-student structure of knowledge distillation. And CutPaste\cite{li2021cutpaste} is a method of one-class classification. In reconstruction-based methods, we selected FAVAE\cite{Dehaene2020AnomalyLB} and DRAEM\cite{zavrtanik2021draem} for our experiments. For comparison purposes, these methods are used as baselines to test the effectiveness of our proposed few-shot IAD.

\subsection{Our Data Augmentation}
It is challenging to select a suitable data augmentation method for unsupervised IAD. Since the training samples are all normal, we cannot use data augmentation to introduce anomalies or make the samples more similar to the abnormal ones. Therefore, some common data augmentation methods such as adding noise, mix-up\cite{zhang2017mixup}, etc. are not applicable. In this paper, we experiment with six data augmentation methods that can potentially enhance the performance of unsupervised IAD. They are rotation, flip, scale, translate, color jitter, and perspective.

\textbf{Rotation.} Rotate the image evenly within [0, 360) degrees. For example, to augment an image to 4 images, we would rotate it by 0, 90, 180, and 240 degrees.

\textbf{Filp.} For an image, the transformation is performed in the order of keeping the original image, flipping horizontally, flipping vertically, and flipping both horizontally and vertically. If we need to generate more pictures, we set the probability of horizontal and vertical flips to 0.5, respectively.

\textbf{Scale.} The side length of the original image is reduced, and the value is uniformly set at (0.5, 1] times the side length of the original image. Enlarging the image may cause the object to be detected beyond the screen, which will cause adverse effects. So only scale the image down a little bit.

\textbf{Translate.} Translate the content of the image horizontally and vertically by [0, 0.5) times the side length of the image. The distance is uniformly set.

\textbf{Color Jitter.} Adjust the brightness and contrast of the image. Except for the original image, the adjustment range is uniformly set at [0.5, 1.5) times, and the original image is always kept at the first image for the use of subsequent mixed data augmentation.

\textbf{Perspective.} It simulates the perspective change of human eyes looking at pictures. The change range is uniform in [0, 0.5).

In this paper, for each augmentation, we augment one training image to four. In other words, in addition to keeping the unmodified image, there are three corresponding training images augmented with this image.

\section{Experimental Results and Analysis}
In this section, we aim to conduct a comprehensive and systematic analysis of the impact of data augmentation on few-shot IAD. We explore several research questions, such as: How does data augmentation affect the performance of different types of IAD algorithms? Can data augmentation methods that achieve better results individually also enhance the accuracy when combined? Do all data augmentation methods exhibit consistent effects across different settings and scenarios? To answer these questions, we use the 6 data augments described in Section 3.3, the 3 datasets introduced in Section 3.1, and the 11 IAD algorithms mentioned in Section 3.2, and thousands of experiments have been conducted.

\textbf{Key Takeaways} (i) There is no universal winner among data augmentation methods for all IAD methods. (ii) We summarize the best single data augmentation for each category of IAD methods in Table~\ref{tab:best_single}. (iii) As for Memory Bank-based methods, we observe that the mixed data augmentation significantly outperforms the single data augmentation for image-level metrics. 

\begin{table*}[htbp]
\setcounter{table}{1} 
\centering
\resizebox{1.95\columnwidth}{!}{%
\begin{tabular}{c|cc|cccc|cc|c|cc}
\hline
\textbf{Classes} &
  \multicolumn{2}{c|}{\textbf{Normalizing Flow}} &
  \multicolumn{4}{c|}{\textbf{Memory Bank}} &
  \multicolumn{2}{c|}{\textbf{Student-Teacher}} &
  \textbf{One-Class} &
  \multicolumn{2}{c}{\textbf{Reconstruction}} \\ \hline
\textbf{Methods} &
  \textbf{CSFlow} &
  \textbf{FastFlow} &
  \textbf{CFA} &
  \textbf{SPADE} &
  \textbf{PaDiM} &
  \textbf{PatchCore} &
  \textbf{RD4AD} &
  \textbf{STPM} &
  \textbf{CutPaste} &
  \textbf{FAVAE} &
  \textbf{DRAEM} \\ \hline
\textbf{Rotation} &
  0.020 &
  0.080 &
  {\color[HTML]{FF0000} 0.007} &
  {\color[HTML]{FF0000} 0.010} &
  {\color[HTML]{FF0000} 0.043} &
  -0.001 &
  {\color[HTML]{333333} 0.036} &
  0.000 &
  {\color[HTML]{333333} 0.067} &
  {\color[HTML]{333333} -0.018} &
  {\color[HTML]{333333} 0.042} \\
\textbf{Flip} &
  0.011 &
  {\color[HTML]{333333} 0.082} &
  {\color[HTML]{FF0000} 0.007} &
  0.006 &
  {\color[HTML]{333333} 0.036} &
  {\color[HTML]{333333} 0.002} &
  {\color[HTML]{FF0000} 0.045} &
  0.008 &
  {\color[HTML]{FF0000} 0.086} &
  {\color[HTML]{333333} 0.000} &
  {\color[HTML]{333333} 0.028} \\
\textbf{Scale} &
  {\color[HTML]{FF0000} 0.045} &
  {\color[HTML]{333333} 0.081} &
  -0.034 &
  0.001 &
  0.001 &
  {\color[HTML]{333333} 0.002} &
  0.021 &
  {\color[HTML]{333333} 0.041} &
  -0.052 &
  -0.056 &
  0.001 \\
\textbf{Translate} &
  {\color[HTML]{333333} 0.037} &
  {\color[HTML]{FF0000} 0.093} &
  -0.012 &
  {\color[HTML]{FF0000} 0.010} &
  0.015 &
  {\color[HTML]{FF0000} 0.007} &
  {\color[HTML]{333333} 0.030} &
  {\color[HTML]{333333} 0.033} &
  -0.030 &
  -0.066 &
  -0.012 \\
\textbf{Color\_Jitter} &
  {\color[HTML]{333333} 0.023} &
  0.066 &
  {\color[HTML]{333333} 0.005} &
  -0.001 &
  {\color[HTML]{333333} 0.017} &
  -0.004 &
  0.022 &
  {\color[HTML]{FF0000} 0.047} &
  -0.035 &
  -0.047 &
  0.012 \\
\textbf{Perspective} &
  -0.006 &
  0.038 &
  -0.001 &
  {\color[HTML]{333333} 0.007} &
  0.014 &
  0.000 &
  0.023 &
  0.030 &
  {\color[HTML]{333333} -0.029} &
  {\color[HTML]{333333} -0.035} &
  {\color[HTML]{FF0000} 0.046} \\ \hline
\end{tabular}%
}
\caption{The average improvement of image-level AUC-ROC of each data augmentation over each IAD method with respect to no data augmentation.}
\label{tab:average}
\end{table*}

\begin{table}[htbp]
\centering
\resizebox{0.6\columnwidth}{!}{%
\begin{tabular}{c|c}
\hline
\textbf{Method}    & \textbf{Best Single Augmentation} \\ \hline
\textbf{CSFlow}    & Scale                             \\
\textbf{FastFlow}  & Translate                         \\
\textbf{CFA}       & Rotation                          \\
\textbf{SPADE}     & Rotation                          \\
\textbf{PaDiM}     & Rotation                          \\
\textbf{PatchCore} & Translate                         \\
\textbf{RD4AD}     & Flip                              \\
\textbf{STMP}      & Color Jitter                      \\
\textbf{CutPaste}  & Flip                              \\ \hline
\end{tabular}%
}
\caption{The best single data augmentation corresponding to various IAD methods.}
\label{tab:best_single}
\end{table}

\begin{figure}[htbp]
  \centering
  \includegraphics[width=0.8\linewidth]{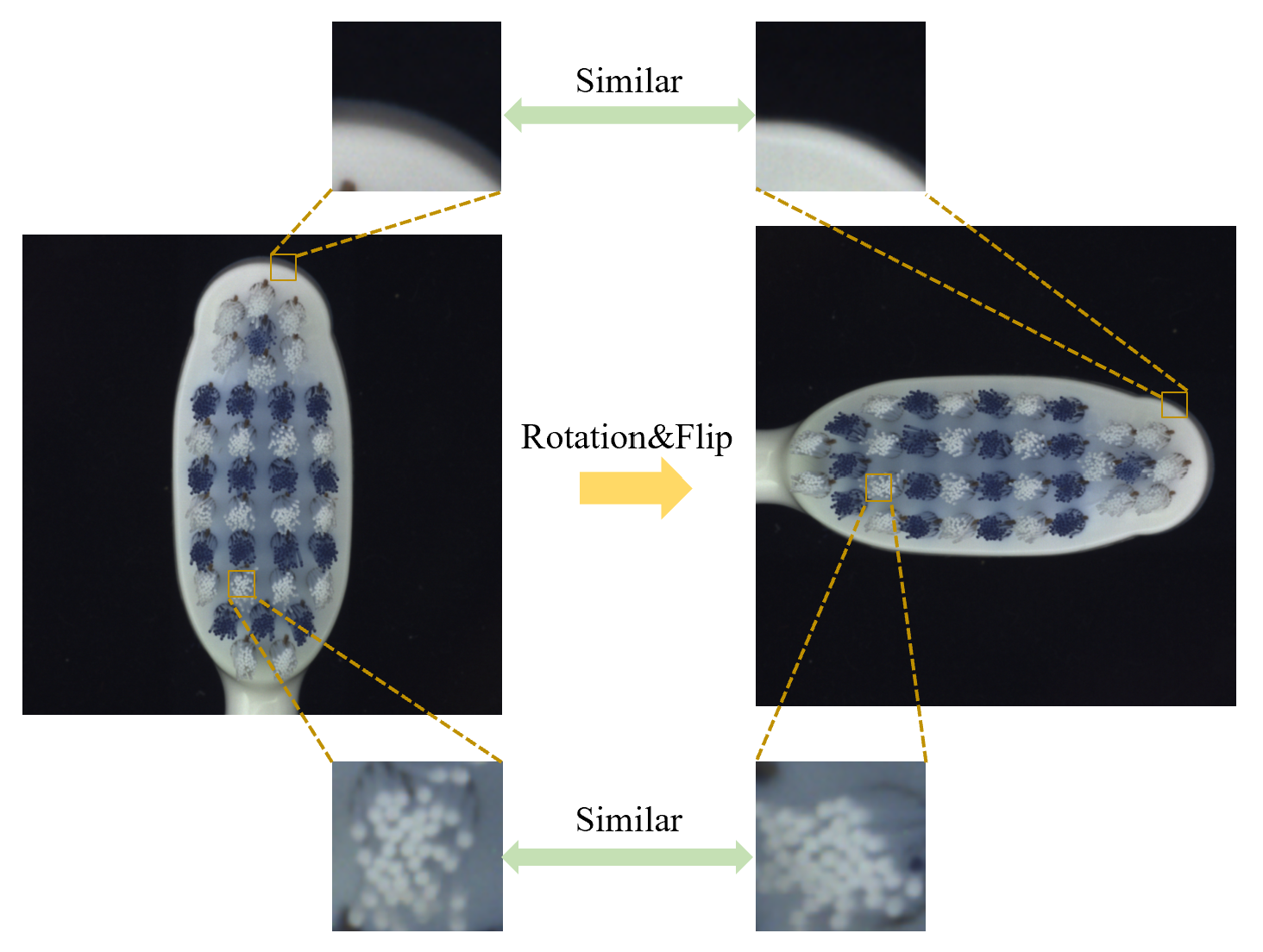}
   \caption{An image from the MvTec AD training set, as well as rotated and flipped images. If you look at the whole image, the data-augmented images are not consistent with the test set. However, if we only look at the patches one by one, the patches after data augmentation can still correspond to the normal toothbrush.}
   \label{fig:toothbrush}
\end{figure}

\begin{table*}[htbp]
\centering
\resizebox{1.5\columnwidth}{!}{%
\begin{tabular}{c|c|cc|cccc|cc|c}
\hline
 &
   &
  \multicolumn{2}{c|}{\textbf{Normalizing Flow}} &
  \multicolumn{4}{c|}{\textbf{Memory Bank}} &
  \multicolumn{2}{c|}{\textbf{S-T}} &
  \textbf{One-class} \\ \hline
\textbf{Shot} &
  \textbf{Dataset} &
  \textbf{CSFlow} &
  \textbf{FastFlow} &
  \textbf{CFA} &
  \textbf{SPADE} &
  \textbf{PaDiM} &
  \textbf{PatchCore} &
  \textbf{RD4AD} &
  \textbf{STPM} &
  \textbf{CutPaste} \\ \hline
 &
  \textbf{MVTec AD} &
  -0.029 &
  {\color[HTML]{FF0000} 0.029} &
  {\color[HTML]{FF0000} 0.006} &
  - &
  {\color[HTML]{FF0000} 0.004} &
  {\color[HTML]{FF0000} 0.004} &
  -0.039 &
  -0.012 &
  {\color[HTML]{FF0000} 0.049} \\
 &
  \textbf{MTD} &
  {\color[HTML]{FF0000} 0.1} &
  -0.026 &
  {\color[HTML]{FF0000} 0.007} &
  - &
  -0.031 &
  {\color[HTML]{FF0000} 0.04} &
  -0.075 &
  -0.025 &
  -0.094 \\
\multirow{-3}{*}{1} &
  \textbf{BTAD} &
  {\color[HTML]{FF0000} 0.001} &
  {\color[HTML]{FF0000} 0.009} &
  -0.004 &
  - &
  0 &
  {\color[HTML]{FF0000} 0.004} &
  -0.227 &
  -0.056 &
  -0.003 \\ \hline
 &
  \textbf{MVTec} AD &
  -0.047 &
  -0.005 &
  -0.005 &
  {\color[HTML]{FF0000} 0.003} &
  -0.007 &
  {\color[HTML]{FF0000} 0.009} &
  -0.087 &
  -0.045 &
  -0.029 \\
 &
  \textbf{MTD} &
  {\color[HTML]{FF0000} 0.011} &
  {\color[HTML]{FF0000} 0.028} &
  -0.001 &
  -0.002 &
  -0.007 &
  {\color[HTML]{FF0000} 0.001} &
  -0.1 &
  {\color[HTML]{FF0000} 0.054} &
  -0.011 \\
\multirow{-3}{*}{2} &
  \textbf{BTAD} &
  -0.026 &
  -0.06 &
  -0.003 &
  0 &
  {\color[HTML]{FF0000} 0.001} &
  {\color[HTML]{FF0000} 0.007} &
  -0.154 &
  -0.035 &
  -0.013 \\ \hline
 &
  \textbf{MVTec AD} &
  -0.123 &
  -0.06 &
  {\color[HTML]{FF0000} 0.002} &
  {\color[HTML]{FF0000} 0.002} &
  0 &
  {\color[HTML]{FF0000} 0.018} &
  -0.104 &
  -0.074 &
  {\color[HTML]{FF0000} 0.052} \\
 &
  \textbf{MTD} &
  {\color[HTML]{FF0000} 0.011} &
  -0.101 &
  {\color[HTML]{FF0000} 0.001} &
  0 &
  {\color[HTML]{FF0000} 0.004} &
  {\color[HTML]{FF0000} 0.023} &
  -0.069 &
  {\color[HTML]{FF0000} 0.028} &
  {\color[HTML]{FF0000} 0.012} \\
\multirow{-3}{*}{4} &
  \textbf{BTAD} &
  -0.029 &
  -0.103 &
  -0.003 &
  {\color[HTML]{FF0000} 0.001} &
  {\color[HTML]{FF0000} 0.002} &
  -0.008 &
  -0.044 &
  -0.073 &
  -0.006 \\ \hline
 &
  \textbf{MVTec AD} &
  -0.21 &
  -0.137 &
  {\color[HTML]{FF0000} 0.007} &
  -0.002 &
  -0.012 &
  {\color[HTML]{FF0000} 0.008} &
  -0.245 &
  -0.142 &
  {\color[HTML]{FF0000} 0.004} \\
 &
  \textbf{MTD} &
  -0.152 &
  -0.08 &
  0 &
  -0.001 &
  {\color[HTML]{FF0000} 0.01} &
  -0.004 &
  -0.133 &
  {\color[HTML]{FF0000} 0.072} &
  {\color[HTML]{FF0000} 0.047} \\
\multirow{-3}{*}{8} &
  \textbf{BTAD} &
  -0.029 &
  -0.15 &
  -0.007 &
  {\color[HTML]{FF0000} 0.002} &
  0 &
  {\color[HTML]{FF0000} 0.024} &
  -0.361 &
  -0.079 &
  -0.036 \\ \hline
\end{tabular}%
}
\caption{The improvement of the mixed data augmentation over the best single data augmentation of the image-level AUC-ROC metric.}
\label{tab:mix}
\end{table*}

\begin{figure*}[htbp]
  \centering
  \includegraphics[width=0.8\linewidth]{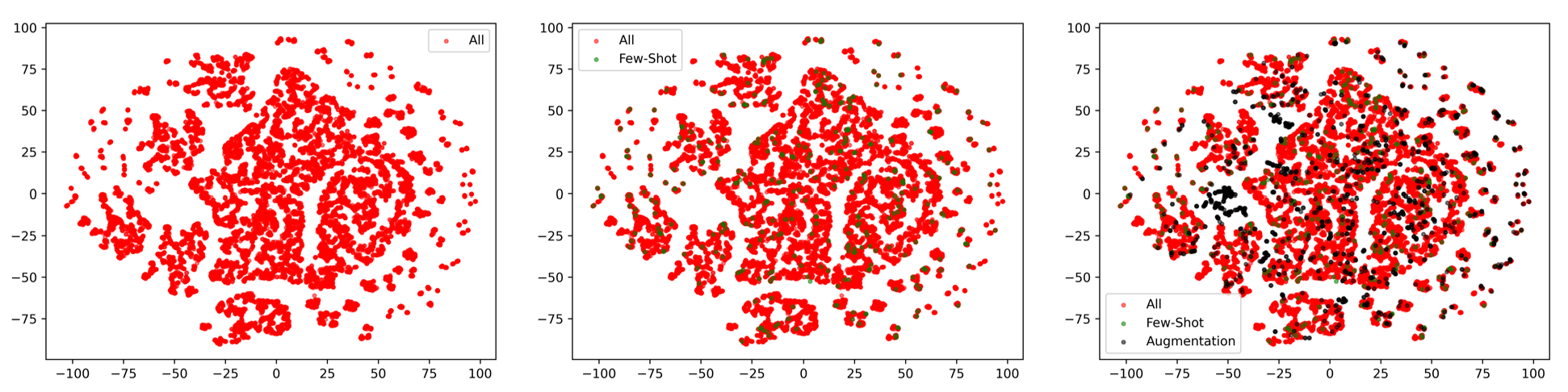}
   \caption{Input all training data of one class of Mvtec AD data set, 4 randomly selected few-shot training images, and images after data augmentation of these four into PatchCore. The distribution of embedding of each patch after dimensionality reduction is shown in the figure.}
   \label{fig:embedding}
\end{figure*}

\subsection{Single Data Augmentation}
As shown in Table \ref{tab:btad}, \ref{tab:mvtec}, \ref{tab:mtd}, a large amount of data is obtained through the experiments. The best and second-best data augmentation for this IAD algorithm in this dataset is marked in red and blue, respectively.

Ideally, we would expect to see a column of the table dominated by either blue or red cells, indicating that the corresponding augmentation method consistently outperforms others. However, this is not the case in our results. Instead, we observe that different augmentation methods have varying effects on different IAD methods. This suggests that there is no single augmentation method that can be universally optimal for all IAD methods. One possible reason for this phenomenon is that different IAD methods have distinct technical designs and mechanisms, which may interact differently with different types of data transformations. Therefore, we further divide the IAD methods into several categories based on their main characteristics and investigate whether similar methods share similar preferences or sensitivities to certain augmentation methods.

As we discussed in Section 3.2, we classify the 11 IAD methods that we evaluate in our experiments into five categories: Normalizing Flow, Memory Bank, Student-Teacher, One-Class, and Reconstruction. These categories are based on the main techniques or principles that each method employs, as shown in Table \ref{tab:average}. For each IAD method and each data augmentation method, we calculate their average improvement of image-level AUC-ROC over all datasets and all shot numbers compared to the baseline of no data augmentation. We observe that within each category of IAD methods, there is some degree of consistency in how data augmentation affects their performance. For example, for Normalizing Flow methods, all data augmentation methods except Perspective can improve their results significantly. For Memory Bank methods, although the overall improvement is modest, Rotation and Flip can consistently boost all four methods in this category. For Student-Teacher methods, all six data augmentation methods have positive effects, especially Flip and Color Jitter. However, for One-Class methods, the impacts of different data augmentation methods vary widely across different models and datasets. Rotation and Flip seem to work well for most models in this category, but other data augmentation methods may have detrimental effects. The only exception is the Reconstruction-based category, where FAVAE\cite{Dehaene2020AnomalyLB} suffers from a decrease in accuracy with any data augmentation method applied while DRAEM\cite{zavrtanik2021draem} benefits from Perspective and Rotation.

\subsection{Mixed Data Augmentation}
We have shown that different data augmentation methods have consistent effects on the performance of IAD methods within the same category. However, we also notice that the optimal data augmentation method may vary depending on the specific model. For instance, for the Student-Teacher category, Flip is the best data augmentation method for RD4AD\cite{deng2022anomaly}, but Color Jitter achieves the highest improvement for STPM\cite{Wang2021StudentTeacherFP}. This motivates us to investigate whether combining two data augmentation methods can further enhance the results compared to using only one. To combine two data augmentation methods, we apply them sequentially to an image. For example, we first rotate an image by 90 degrees and then translate it by a certain amount. In this way, we can generate more diverse and complex variations of the original image.

We choose different combinations of data augmentation methods for different categories of IAD methods based on their performance in the previous experiments. For Normalizing Flow-based methods, we combine Translate and Scale, which are both geometric transformations that can preserve the shape and appearance of the objects in the image. For Memory Bank-based and One-Class-based methods, we combine Rotation and Flip, which are both simple and effective data augmentation methods that can introduce some degree of rotation invariance and symmetry to the models. For Student-Teacher-based methods, we combine Translate, Color Jitter, and Flip, which can generate more diverse and challenging variations of the image by changing its position, color, and orientation. We do not conduct experiments with mixed data augmentation for Reconstruction-based methods because we have observed that data augmentation does not improve their performance and may even degrade it. This may be because data augmentation interferes with the reconstruction objective of these methods and makes it harder for them to learn meaningful features from the images.

We take the best single data augmentation corresponding to various IAD methods as the baseline for mixed data augmentation. As shown in Table \ref{tab:best_single}, the best single data augmentation corresponding to each method varies. Next, we will investigate whether mixed data augmentation can further improve accuracy.

Table \ref{tab:mix} demonstrates the improvement of the mixed data augmentation over the best single data augmentation in the image-level AUC-ROC metric. One can see that mixed data augmentation only has a more obvious improvement on PatchCore\cite{roth2022towards}, and even causes a decline in accuracy in other methods.

\begin{figure}[htbp]
  \centering
  \includegraphics[width=0.77\linewidth]{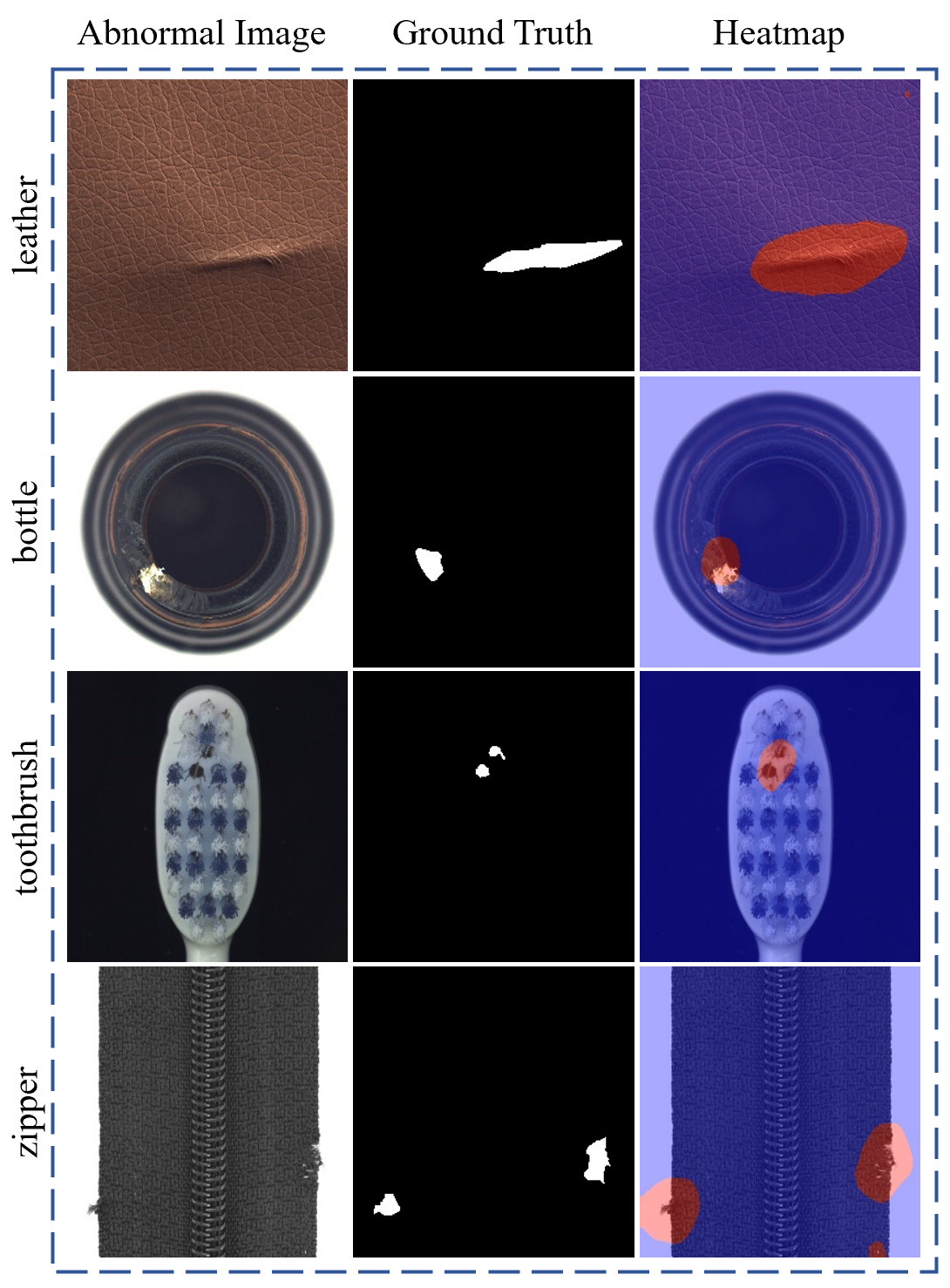}
   \caption{Using PatchCore to perform mixed data augmentation of 4-shot IAD in the Mvtec AD dataset can get good results in all classes.}
   \label{fig:vis}
\end{figure}

\subsection{Analysis}
The main purpose of applying data augmentation to the training data is to increase its diversity and make it more similar to the distribution of the test data. However, as we can see from Figure \ref{fig:toothbrush}, some data augmentation methods may change the orientation or appearance of the objects in the training images in ways that do not match the test images. For example, after applying Rotation and Flip to a toothbrush image in the training set, the toothbrush may appear upside down or sideways, which is unlikely to occur in the test set. Therefore, for most IAD methods, using multiple data augmentation methods together may not be beneficial and may even harm their performance. However, there are two exceptions: PaDiM\cite{defard2021padim} and PatchCore\cite{roth2022towards}. These two methods do not rely on the global shape or structure of the image but rather on its local patches, as shown in Figure \ref{fig:toothbrush}. Even if the image is transformed by data augmentation, each patch still retains some similarity to the normal image patches. However, PaDiM performs Gaussian fitting for patches at the same spatial location across different images and expects them to have low variance. Data augmentation introduces more diversity and noise to these patches and thus degrades PaDiM’s performance. On the other hand, PatchCore does not care about the spatial location or global changes of the patches and only focuses on their local features. Therefore, data augmentation has little negative impact on PatchCore’s performance and may even improve it further by generating more challenging variations of normal images.

To gain more insight into how PatchCore\cite{roth2022towards} works and to verify our previous claim, we conduct an experiment using the Mvtec AD dataset. We feed all the training images of one class, four randomly selected few-shot training images, and their augmented versions into PatchCore and extract the embeddings of all the patches. We then use t-SNE\cite{van2008visualizing} to reduce the dimensionality of all the embeddings to two and visualize them in Figure \ref{fig:embedding}. We can observe that the embeddings from the few-shot images are very sparse and only cover a small region of the feature space spanned by the full training data. However, after applying data augmentation to these images, their embeddings become dense and cover a larger region of the feature space. This suggests that data augmentation can effectively generate more diverse and realistic patches and thus improve the performance of few-shot IAD with PatchCore.

Figure \ref{fig:vis} shows some examples of PatchCore’s\cite{roth2022towards} performance on different objects with mixed data augmentation using only 4-shot training images. We can see that PatchCore can not only correctly classify the images as normal or abnormal but also precisely locate the abnormal regions on the objects. This is because the patch embeddings obtained from data augmentation are similar to those from normal images and form a tight cluster in the feature space. Therefore, PatchCore can easily distinguish them from the abnormal patches that deviate from this cluster. As a result, PatchCore does not suffer from false positives, which means that it does not mistakenly label normal regions as abnormal.

\section{Conclusions}
In this paper, we conduct a comprehensive study on the role of data augmentation in few-shot IAD. We evaluate 11 algorithms on 3 datasets using 6 types of data augmentation methods and analyze their performance and characteristics. The experimental results show that data augmentation has different effects on different IAD algorithms depending on their underlying principles and assumptions. However, for algorithms that share similar technical approaches, such as embedding-based methods, data augmentation plays a similar role. Moreover, we also discover that while some data augmentation methods may improve the performance of a certain algorithm individually, combining them together may not be beneficial and may even degrade the anomaly detection accuracy. This is because some data augmentation methods may introduce unrealistic or inconsistent variations to the images that do not match the test images. However, PatchCore is an exception to this rule because it only focuses on the local features of patches and ignores their spatial location or global changes. Mixed data augmentation does not affect its accuracy but rather enhances it by generating more diverse and challenging normal images.

\section{Acknowledgments} 
This work is supported by the National Natural Science Foundation of China under Grant No. 62122035, 62206122, and 61972188.

{\small
\bibliographystyle{ieee_fullname}
\bibliography{egbib}
}

\end{document}